\documentclass[letterpaper, 10 pt, conference]{IEEEconf}  
\IEEEoverridecommandlockouts
 
\usepackage{cite}
\usepackage{amsmath,amssymb,amsfonts}
\usepackage{graphicx}
\usepackage{textcomp}
\usepackage{color}
\usepackage{subfigure}
\usepackage{multirow}
\usepackage{multicol}
\usepackage{makecell}
\usepackage{bm}
\usepackage{algorithm}
\usepackage{algpseudocode}
\usepackage{hyperref}
\newcommand{\ie}{i.e.,\ }
\newcommand{\eg}{e.g.,\ }
\newcommand{\Reffig}[1]{Figure~\ref{#1}}
\newcommand{\Refsec}[1]{Section~\ref{#1}}

\newcommand{\Refeq}[1]{Equation~\eqref{#1}}
\newcommand{\Reftab}[1]{Table~\ref{#1}}

\hypersetup{hidelinks,
        colorlinks=true,
        allcolors=blue,
        pdfstartview=Fit,
        breaklinks=true}
\usepackage{graphics} 
\usepackage{epsfig} 
\usepackage{algorithm}
\usepackage{algpseudocode}
\usepackage{amsmath}
\usepackage{amssymb}
\usepackage{subfigure}
\usepackage{graphicx}
\usepackage{subfigure}
\usepackage{booktabs}
\usepackage{threeparttable}
\usepackage{multirow}
\usepackage{multicol}
\usepackage{makecell}
\usepackage{dcolumn}
\usepackage[T1]{fontenc}
\usepackage{diagbox}

\begin{document}
\title{
        LOG-LIO: A LiDAR-Inertial Odometry with Efficient Local Geometric Information Estimation
}

\author{Kai Huang$^{1}$, Junqiao Zhao$^{*,2, 3}$, Zhongyang Zhu$^{2, 3}$, Chen Ye$^{2}$, Tiantian Feng$^{1}$ 
\thanks{$^{1}$Kai Huang and Tiantian Feng are with the School of Surveying and Geo-Informatics, Tongji University, Shanghai, China
{\tt\footnotesize (e-mail: huangkai@tongji.edu.cn; fengtiantian@tongji.edu.cn).}}
\thanks{$^{2}$Junqiao Zhao, Zhongyang Zhu and Chen Ye are with Department of Computer Science and Technology, School of Electronics and Information Engineering, Tongji University, Shanghai, China, and the MOE Key Lab of Embedded System and Service Computing, Tongji University, Shanghai, China
{\tt\footnotesize (e-mail: zhaojunqiao@tongji.edu.cn; 2233057@tongji.edu.cn; yechen@tongji.edu.cn).}}
\thanks{$^{3}$Institute of Intelligent Vehicles, Tongji University, Shanghai, China}
\thanks{This work is supported by the National Key Research and Development Program of China (No. 2021YFB2501104). \emph{(Corresponding Author: Junqiao Zhao.)}}
}

\maketitle

\begin{abstract}

        Local geometric information, \ie normal and distribution of points, is crucial for LiDAR-based simultaneous localization and mapping (SLAM) because it provides constraints for data association, which further determines the direction of optimization and ultimately affects the accuracy of localization.
        However, estimating normal and distribution of points are time-consuming tasks even with the assistance of kdtree or volumetric maps.
        To achieve fast normal estimation, we look into the structure of LiDAR scan and propose a ring-based fast approximate least squares (Ring FALS) method.
        With the Ring structural information, estimating the normal requires only the range information of the points when a new scan arrives.
        To efficiently estimate the distribution of points, we extend the ikd-tree to manage the map in voxels and update the distribution of points in each voxel incrementally while maintaining its consistency with the normal estimation.
        We further fix the distribution after its convergence to balance the time consumption and the correctness of representation.
        Based on the extracted and maintained local geometric information, we devise a robust and accurate hierarchical data association scheme where
        point-to-surfel association is prioritized over point-to-plane.
        Extensive experiments on diverse public datasets demonstrate the advantages of our system compared to other state-of-the-art methods.
        Our open source implementation is available at \href{https://github.com/tiev-tongji/LOG-LIO}{https://github.com/tiev-tongji/LOG-LIO}.
\end{abstract}


\section{INTRODUCTION}
\label{sec:introduction} 


The performance of LiDAR (-inertial)-based simultaneous localization and mapping (SLAM) system depends heavily on the registration between the LiDAR scan and the map, \ie{finding the correspondence between them based on the similarity of the local geometric information and then minimizing their distance}.
Local geometric information includes attributes that can represent the position, shape, and other characteristics of the local surface where a point is located.

The conventional method for estimating local geometric information is to evaluate the smoothness of the input scan and to locally approximate the map with geometric primitives \cite{zhang2014loam}.
However,
accurate estimation of local geometric information requires the retrieval of neighborhood information in a dense point cloud, but for LiDAR-inertial odometry (LIO) systems, this results in a huge computational burden even with the help of kdtree or volumetric maps.

Accurate and fast estimation of local geometric information has gained increasing attention in recent studies \cite{chen2022direct, nguyen2023slict, reinke2022locus2, palieri2020locus, ramezani2022wildcat}.
Among them, the normal and the distribution of points are two representative attributes, since the former indicates the tangent plane of the local surface, and the latter implies the average position and shape of the point cloud sampled from the local surface.
However, current LIO systems seldom incorporate the real-time estimation of the normal and the distribution of points, which hampers their pose estimation performance.

This paper presents LOG-LIO, a robust and accurate LIO system focusing on the real-time estimation of the normal of LiDAR scan points and the distribution of map points, and their rational utilization.
Inspired by \cite{badino2011fast} and \cite{fan2021three}, we look into the structure of a LiDAR scan and propose a Ring-based fast approximate least squares method, namely Ring FALS. 
We project point cloud onto the range image to pre-build a lookup table, which represents the structural information of the specific LiDAR.
With the arrival of a new scan, only the range information of the points is needed to estimate the normal.
We incrementally update the distribution of points for each voxel in the map while maintaining its consistency with the normals.
To balance time consumption and correctness of representation, we manage the map on the extended ikd-tree and further fix the distribution after it converges.

Similar to the FAST-LIO series \cite{xu2021fast, xu2022fast}, we directly associate scan points with voxels on the map after distortion correction.
For scan points that satisfy visibility and consistency checks based on normals, we devise a robust and accurate hierarchical data association scheme considering the distribution.
The poses are optimized by integrating the IMU measurements as initial estimates and then using an error-state iterative Extended Kalman filter (iEKF) \cite{xu2021fast} to minimize the multi-scale point-to-surfel and point-to-plane distances.

The main contributions of this work are as follows:
\begin{itemize}
        \item
              Ring FALS, modified from FALS, a normal estimator that utilizes the structural information of LiDAR scan can meet the real-time requirements of the LIO system.
              \item
              A robust and accurate hierarchical data association scheme considering the distribution of points within map voxels where point-to-surfel is prioritized over point-to-plane and large-scale surfel over small-scale surfel.
        \item
              Extensive experiments on public datasets demonstrate the advantages of our LIO system compared to other state-of-the-art methods.
              To benefit the community, our implementation of this work is open-source at \href{https://github.com/tiev-tongji/LOG-LIO}{https://github.com/tiev-tongji/LOG-LIO}, and we also open-source Ring FALS as an independent normal estimation tool at \href{https://github.com/tiev-tongji/RingFalsNormal}{https://github.com/tiev-tongji/RingFalsNormal}.
\end{itemize}


\section[sec:related_work]{RELATED WORKS} 
\label{sec:related_work}

\subsection{Point Cloud Normal Estimation}


The most commonly used method to obtain surface normals from point cloud is the least square estimation based on the neighborhood search due to its ease of implementation \cite{rusu20113d}.
However, the least squares-based method is computationally expensive for LIO systems.


\cite{badino2011fast} compares the complexity of least squares approaches and proposes FALS, which simplifies the least squares loss function to completely avoid the computation of the covariance matrix for each point.
\cite{badino2011fast} also reformulates the traditional least squares solution to estimate the normal by calculating the derivatives of the surface from a spherical range image (SRI).
However, the number of multiplications for normals computation of SRI is greater than FALS.

Rather than least square-based solution, 3F2N\cite{fan2021three} performs three filtering operations on the inverse depth image to estimate the normals, which has the comparative performance as FALS, but its efficiency and accuracy are strongly influenced by the filter selection.


Inspired by \cite{badino2011fast} and \cite{fan2021three}, we propose Ring FALS. 
We pre-build a lookup table which represent the structural information for the specific LiDAR.
Compared to FALS, Ring FALS further simplifies the projection of each LiDAR scan with the assistance of ring index, while preserving accuracy.

\subsection{Distribution of Points Estimation}

The distribution of a point is represented through its 3D coordinates and the covariance matrix computed by its neighboring points.
\cite{segal2009generalized} proposes the generalized ICP (GICP) algorithm, which takes into account the locally planar structure of points in a probabilistic model and then minimizes the distance between distributions.
But searching neighboring points to compute the covariance matrix is too time-consuming for LIO systems.

LOAM \cite{zhang2014loam} does not estimate the distribution of points, but performs Eigen analysis on the associated map points to determine whether its local geometry is a line or a plane.
However, the coordinates of the sparse map points cannot accurately represent the local geometric information, which leads to inaccurate constraints for the registration.

DLO \cite{chen2022direct} registers point cloud using GICP to minimize the plane-to-plane distance, which is derived from the covariance matrix of each point.
It assumes that the covariance of submap can be approximated by concatenating the normals from keyframes, and the covariance of points is only computed once when the scan is acquired.
However, such a normals stitching method cannot accurately reflect the local geometric information of the point cloud, which ultimately affects accuracy.
LOCUS 2.0 \cite{reinke2022locus2} extends the work of LOCUS \cite{palieri2020locus}, which constructs covariance matrices for GICP-based registration based on the pre-computed normals, but how to pre-compute normals is not elaborated in their paper.

Wildcat \cite{ramezani2022wildcat} fits ellipsoids based on the coordinates and timestamps of the clustered points. 
The ellipsoids representing the distribution of points are further used to generate surfels.
SLICT \cite{nguyen2023slict} further proposes an octree-based global map and updates the distribution of points within each voxel incrementally.
It obtains large-scale distributions by merging multiple voxels to generate surfels in multi-resolution.

Inspired by the above methods, we extend ikd-tree to maintain the distribution of points in each map node incrementally, and fix the distribution after its convergences.

\subsection{LiDAR (-Inertial) Odometry}

LOAM \cite{zhang2014loam} has inspired many LiDAR SLAM systems due to the low coupling of system modules and the rational use of point cloud geometry attributes.
However, the lack of effective map management and the high time consumption required for optimization can degrade the performance of the system. 

LIO-SAM \cite{kaess2012isam2} proposes a framework based on keyframes and local maps, which optimizes poses in a factor graph.
However, LIO-SAM builds sub-maps for input scans by simply merging point cloud of surrounding keyframes, 
which is a time-consuming process when the number of points is large compared to incrementally maintaining maps.

FAST-LIO \cite{xu2021fast} employs point-to-plane correspondence and the iEKF to directly register the LiDAR scan and the map.
It presents a new formula to compute the Kalman gain, and the computation load only depends on the dimension of state dimension.
FAST-LIO2 \cite{xu2022fast} maintains the map by an incremental kdtree data structure, namely ikd-tree, to further improve efficiency.

In this paper, we adopt the iEKF to optimize poses by directly registering the LiDAR scan and the map, but with a different data association scheme.
By incorporating real-time normal and distribution of points estimation, we can efficiently construct surfels in multi-scale, which represent the local surface geometry more accurately than other geometric primitives, \eg plane.
We prioritize associating large-scale surfels over small-scale surfels since large-scale surfels are modeled with more points and are insensitive to noise.

\section{PRELIMINARY}
\label{sec:preliminary}

\subsection{Notation}
\label{sec:notation}
We now define notations and frames that we used throughout the paper.
We consider $\mathcal{W} $ as the world frame and ${\mathcal{I} _k}$, ${\mathcal{L} _k}$ as the IMU and LiDAR frames, related to the $k$-th LiDAR scan at time $t_k$, respectively.
$^a\mathbf{T}_b \in SE(3)$ to be Euclidean transformation take 3D points from frame $b$ to frame $a$, which is consisted of rotation $^a\mathbf{R}_b \in SO(3)$  and translation $^a\mathbf{t}_b \in \mathbb{R}^3$.
$\boldsymbol{n}_r$ denotes the normal from Ring FALS and 
$\boldsymbol{e}_d$ is the eigenvector corresponding to the smallest eigenvalue of a distribution (see \Refsec{sec:surfel}).

\subsection{LiDAR Observation Model}
\label{sec:lidar_obser}
\begin{figure}[!ht]
        \centering
        \subfigure[LiDAR observation model]
        {
                \begin{minipage}[b]{0.2\textwidth}
                        \includegraphics[width=1\textwidth]{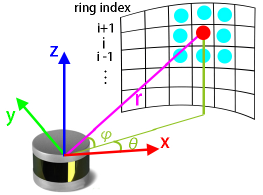}
                \end{minipage}
                \label{fig_lidar_obser}

        }
        \subfigure[multi-scale surfel association]
        {
                \begin{minipage}[b]{0.2\textwidth}
                        \includegraphics[width=1\textwidth]{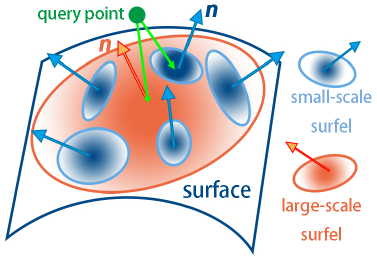}
                \end{minipage}
                \label{fig_association}
        }
        \caption{Illustration of the LiDAR observation model and multi-scale surfel association.
                (a) 
                The magenta line indicates the ray of the red point.
                The eight points are the neighborhoods that Ring FALS uses to estimate the normal of the red point.
                (b) 
                The orange ellipses represent the large-scale surfel merged by the five blue small-scale surfels.   
                }
\end{figure}

In practice, LiDAR obtains the 3D coordinates of a point by combining bearing and range measurements of the target surface\cite{yuan2021pixel, yuan2022efficient}, as shown in \Reffig{fig_lidar_obser}.
The LiDAR observation model is as follows:
\begin{equation}
        \boldsymbol{p}_i = r_i\boldsymbol{v}_i = r_i \left[\begin{array}{c}
                        \cos \theta_i \cos \varphi_i \\
                        \sin \theta_i \cos \varphi_i \\
                        \sin \varphi_i
                \end{array}\right]
        \label{eq_p_rv}
\end{equation}
where $r_i$ is the range, $\theta_i$ the azimuth and $\varphi_i$ the vertical angle of the target point.
$\boldsymbol{v}_i$ represents the horizontal and vertical structural information of the point relative to the LiDAR.

For a spinning LiDAR, 
we denote the horizontal resolution as $H_{res}= 2\pi/m$, where $m$ is the constant number of points within each ring. 
Denoting the structural information $\boldsymbol{s}_i = \left[\cos \theta_i\ \sin \theta_i\ \cos \varphi_i\ \sin \varphi_i \right]^T$
and then $\boldsymbol{s}_i$ can be arranged into a lookup table $\mathcal{T}$ based on the ring index and azimuth relative to the LiDAR as follows:
\begin{equation}
        \boldsymbol{s}_i = \mathcal{T}  (row_i,\ col_i) 
        \label{eq_p_v_rc}
\end{equation}
where $row_i$ represents the ring index of $\boldsymbol{p}_i$ and $col_i = round(\theta_i / H_{res}) $.

\subsection{Least Squares Normal Estimation}
\label{sec:ls_ne}
Given a subset of $n$ 3D points $\boldsymbol{p}_i$, $i = 1,2,...,n$ of the surface, least squares finds the normal vector $\boldsymbol{n}=(n_x,n_y,n_z)$ and the scalar $d$ that minimizes \Refeq{eq_pn_d}.
\begin{equation}
        e = \sum_{i = 1}^{n}(\boldsymbol{p}_i^T \boldsymbol{n} - d)^2
        \label{eq_pn_d}
\end{equation}
The closed form solution of the normal $\boldsymbol{n}$ is the eigenvector corresponding to the smallest eigenvalue of the covariance matrix in \Refeq{eq_M}.
\begin{equation}
        \boldsymbol{M} = \sum_{i = 1}^{n}(\boldsymbol{p}_i - \overline{\boldsymbol{p}})(\boldsymbol{p}_i - \overline{\boldsymbol{p}})^T
        \label{eq_M}
\end{equation}
with $\overline{\boldsymbol{p}} = 1/n\sum_{i = 1}^{n}\boldsymbol{p}_i$.

\subsection{Distribution of Points}
\label{sec:distribution_estimation}


The distribution of points within a voxel can be represented by its mean position $\overline{\boldsymbol{p}}$ and the covariance matrix $\boldsymbol{M}$, as shown in \Refeq{eq_M}.
And $\boldsymbol{M}$ can be further simplified as follows:
\begin{equation}
        \boldsymbol{M} = \mathcal{S}_n - \frac{1}{n} \mathcal{P}_n \mathcal{P}_n^T
        \label{eq_M_simp}
\end{equation}
where $\mathcal{S}_n $ denotes $\sum_{i = 1}^{n}\boldsymbol{p}_i \boldsymbol{p}_i^T$ and $\mathcal{P}_n$ denotes $\sum_{i = 1}^{n}\boldsymbol{p}_i $.
Due to the symmetric nature of $\boldsymbol{M}$, it is only necessary to record the six elements in its upper right corner.

The accurate representation of the distribution of points requires a large number of points.
Due to limited resolution and occlusion, point cloud from multiple locations must be accumulated incrementally to obtain high quality maps.
For newly incorporated $m$ points in a voxel, their $\mathcal{S}_m$, $\mathcal{P}_m$ need to be calculated.
Subsequently, the distribution of points within this voxel can be updated by \cite{thrun2002probabilistic}:
\begin{equation}
        \begin{array}{c}
                \overline{\boldsymbol{p}} = (\mathcal{P}_n + \mathcal{P}_m) / (n + m)
                \\
                \boldsymbol{M} = \mathcal{S}_n + \mathcal{S}_m - \frac{1}{n + m}(\mathcal{P}_n + \mathcal{P}_m)(\mathcal{P}_n + \mathcal{P}_m)^T
        \end{array}
        \label{eq_M_merge}
\end{equation}

\subsection{Surfel}
\label{sec:surfel}
We define the planarity $\rho$ within a voxel
similar to SLICT \cite{nguyen2023slict}, and further introduce $\gamma$ as following:
\begin{equation}
        \begin{array}{c}
                \rho_i = 2 (\lambda_2 - \lambda_1) / (\lambda_1 + \lambda_2 + \lambda_3) \\
                \gamma_i =  \displaystyle{\lambda_2 / \lambda_1}
        \end{array}
        \label{eq_planarity}
\end{equation}
where $\lambda_1, \lambda_2, \lambda_3$ are the eigenvalues of covariance matrix $\boldsymbol{M}$ with $\lambda_1 < \lambda_2 < \lambda_3$.
We define a surfel has $\rho_i$ greater than 1.0 and $\gamma_i$ greater than 100.
A larger $\rho_i$ implies that the distribution of the sampled points is flatter on the surface, and a larger $\gamma_i$ indicates that the distribution is less close to a linear geometry.
If the above criteria are satisfied, the surfel is represented by the mean position of points $\overline{\boldsymbol{p}}$ and the normal $\boldsymbol{e}_d$, where $\boldsymbol{e}_d$ is the eigenvector corresponding to $\lambda_1$.
Multiple small-scale surfels can be merged into a large-scale surfel by merging the distributions following \Refeq{eq_M_merge}.
And the merged distribution still needs to satisfy the criteria in the above to be considered as a large-scale surfel.

\section{Ring FALS Normal Estimator}
\label{sec:ring_fals}

We first revisit FALS \cite{badino2011fast}.
In FALS, \Refeq{eq_pn_d} is reformulated to obtain:
\begin{equation}
        \widetilde{e} = \sum_{i = 1}^{n}(\boldsymbol{p}_i^T \widetilde{\boldsymbol{n}} - 1)^2
        \label{eq_pn_1}
\end{equation}
where $ \widetilde{\boldsymbol{n}}$ is defined up to a scale factor.
Substituting \Refeq{eq_p_rv} gives:
\begin{equation}
        \widetilde{e} = \sum_{i = 1}^{n}r_i^2(\boldsymbol{v}_i^T \widetilde{\boldsymbol{n}} - r_i^{-1})^2
        \label{eq_rvn_r}
\end{equation}
where $r_i$ is the range and $\boldsymbol{v}_i$ implies the bearing information of the target points related to the LiDAR.



It can be assumed that the range of points within a small region are similar, thanks to the high-resolution LiDAR.
Therefore, $r_i^2$ can be removed from \Refeq{eq_rvn_r} to obtain an approximation:
\begin{equation}
        \widehat{e} = \sum_{i = 1}^{n}(\boldsymbol{v}_i^T \widehat{\boldsymbol{n}} - r_i^{-1})^2
        \label{eq_vn_r}
\end{equation}
where $\widehat{\boldsymbol{n}}$ is the approximate normal, and it has the closed form solution $\widehat{\boldsymbol{n}} = \widehat{\boldsymbol{M}}^{-1}\widehat{\boldsymbol{b}}$ where $\widehat{\boldsymbol{M}} =  \sum_{i = 1}^{n}\boldsymbol{v}_i\boldsymbol{v}_i^T$ and $\widehat{\boldsymbol{b}} = \sum_{i = 1}^{n}\boldsymbol{v}_i/r_i$.
The matrix $\widehat{\boldsymbol{M}}^{-1}$ depends only on the constant structural information $\boldsymbol{v}$, independent of the range $\boldsymbol{r}$.
Hence, the matrix $\widehat{\boldsymbol{M}}^{-1}$ can be pre-computed as a lookup table. 

To obtain the neighborhood of $\boldsymbol{v}_i$ for computing $\widehat{\boldsymbol{M}}^{-1}$ and $\widehat{\boldsymbol{b}}$, 
FALS projects the LiDAR scan points onto an SRI following \Refeq{eq_p_rv}. 
The computation of $\widehat{\boldsymbol{M}}^{-1}$ and $\widehat{\boldsymbol{b}}$ requires that each pixel in the SRI has a corresponding measurement, hence an interpolation is needed since the LiDAR scan points only occupy sparse pixels.
This is in turn a time-consuming process.

Different from FALS, Ring FALS establishes a fast mapping following \Refeq{eq_p_v_rc} based on the structural information of the LiDAR. 
To avoid the costly vacant pixel interpolation, we create a table with the number of rows corresponding to the number of rings and the number of columns matching the number of points within each ring. 
The table is created solely based on the provided measurements, eliminating the necessity for interpolation.
Thus Ring FALS speeds up the projection step in FALS and circumvents the time-consuming neighborhood search in many LIO systems, facilitating dense normal estimation for LiDAR scans.



Note that there are instances where the assumption of Ring FALS may not hold.
Such cases include scenarios like wall edges, occlusions where the range of points varies significantly in a small area, and situations with missing range measurements.
To address this, we flip all the backfaced normals and then apply image median blurring to smooth the normals and enhance their consistency.
For points whose normal direction still differs significantly from the associated map points, we identify them as outliers in the optimization process through visibility and consistency checks, as elaborated in \Refsec{sec:consistency_checks}.



\section{System Description}
\label{sec:system_description}

\begin{figure*}[!ht]
        \centering
        \includegraphics[width=16cm]{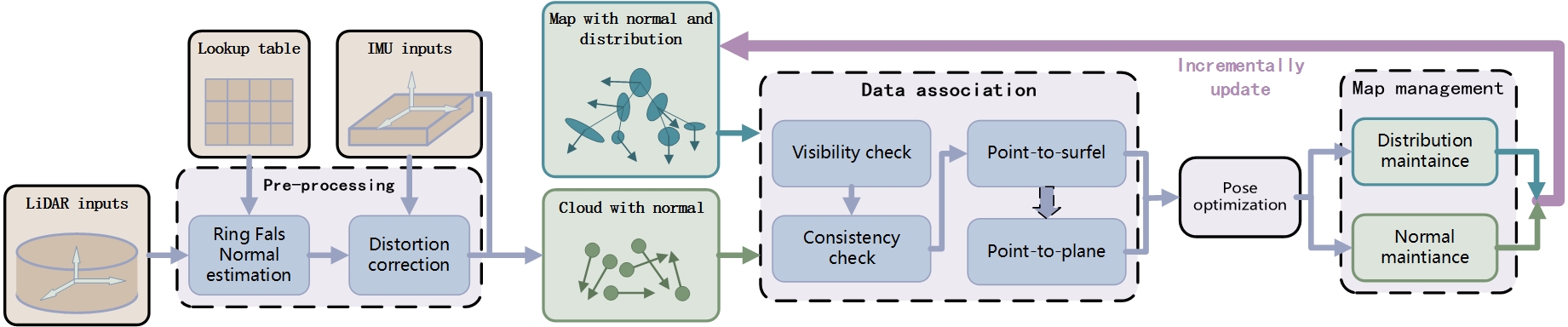}
        \caption{System overview of LOG-LIO}
        \label{fig_overview}
\end{figure*}
The pipeline of LOG-LIO is shown in \Reffig{fig_overview}.
For a new input scan, we first estimate the normal of the points.
The association is then performed between the undistorted point cloud and the map according to their local geometric information.
We incorporate the measurements of IMU and optimize the poses of the body via iEKF.
After optimization, new points are added to the map managed by the extended ikd-tree, and the distribution within a voxel is incrementally maintained. 

Our system takes the IMU frame as the body frame, where the system state $\mathbf{x}$ can be written as:
\begin{equation}
        \mathbf{x} = \left[^\mathcal{W}\mathbf{R}^T_\mathcal{I}\ ^\mathcal{W}\mathbf{p}^T_\mathcal{I}\ ^\mathcal{W}\mathbf{v}^T_\mathcal{I}\ \mathbf{b}^T_\omega\ \mathbf{b}^T_a\ ^\mathcal{W}\mathbf{g}^T\right]
        \label{eq_x}
\end{equation}
where$^\mathcal{W}\mathbf{R}^T_\mathcal{I}$, $^\mathcal{W}\mathbf{p}^T_\mathcal{I}$ and $^\mathcal{W}\mathbf{v}^T_\mathcal{I}$ are the orientation, position and velocity of IMU in the world frame (\ie the first IMU frame), $\mathbf{b}^T_\omega$ and $\mathbf{b}^T_a$ are gyroscope and accelerometer bias respectively, $^\mathcal{W}\mathbf{g}^T$ is the known gravity vector in the world frame.

\subsection{Data Pre-processing}
\label{sec:data_pre}

LOG-LIO uses Ring FALS (\Refsec{sec:ring_fals}) to to estimate the normal for each input point, denoted as $\boldsymbol{n}_r$.
Subsequently, voxel grid downsampling and backward propagation based on IMU measurements are used for point reduction and distortion correction, respectively.

\subsection{Data Association}
\label{sec:data_association}

At the beginning of data association, the IMU measurements are integrated from the previous frame to predict the pose $\widehat{\mathbf{x}}_{k}$.
Using this prediction, each new input point $^\mathcal{L}\boldsymbol{p}_i$ is transformed to the world frame $^\mathcal{W}\boldsymbol{p}_i = ^\mathcal{W}\widehat{\boldsymbol{T}}_\mathcal{I}\  ^\mathcal{I}\boldsymbol{T}_\mathcal{L}\  ^\mathcal{L}\boldsymbol{p}_i$.
Then, data association is performed in three consecutive steps:

\subsubsection{Initial Correspondence}
For a query point $^\mathcal{W}\boldsymbol{p}_i$, we first search for its $k$ nearest map points corresponding to $k$ nearest map voxels.

\subsubsection{Visibility and Consistency Checks}
\label{sec:consistency_checks}
The candidate associated map points may not be visible to the LiDAR if the angle between the normal of the map point and the ray (vector from the query scan point to the LiDAR center) is greater than 90 degrees.
Such a case usually occurs indoors, where the two planes of an object (\eg a wall) are close to each other, which is referred to as the double-side issue\cite{zhou2021lidar}.
This incorrect correspondence is eliminated directly.

The consistency of the associated map points is evaluated by computing the average angle between the normal of the query point and the normals of the associated map points.
If the average angle is larger than a threshold $\alpha$ ($\alpha = 60^\circ$), we consider it an inconsistent association and discard it.

\subsubsection{Hierarchical Association}
\label{sec:hierarchical}

For query points satisfying visibility and consistency checks, a hierarchical association is performed, 
where point-to-surfel is prioritized over point-to-plane, and large-scale surfels are prioritized over small-scale surfels.

Surfels offer a more precise and flexible representation of a local surface compared to a plane fitted with sparse map points since they are modeled with the distribution of points, 
which not only indicates the location but also captures the shape of the local surface. 
Large-scale surfel can be approximated by merging multiple small-scale distributions following \Refeq{eq_M_merge}.
Moreover, they exhibit a high tolerance to noise, thereby providing more robust constraints when contrasted with small-scale surfels.
As illustrated in \Reffig{fig_association}, the orange ellipse depicts the large-scale surfel merged by the five blue small-scale surfels. 
The green query point is initially associated with the merged large-scale surfel if the merged large-scale surfel satisfies the criteria outlined in \Refsec{sec:surfel}, 
and the distance from the mean position of each small-scale surfel to the large-scale surfel is below a predefined threshold. 
Otherwise, the association with small-scale surfels is preferred.

For constraints with small-scale surfels, we associate the query point with the surfel of the voxel where the point is located, which must already be fixed.
If the voxel cannot meet the criteria of a surfel (\Refsec{sec:surfel}), we resort to using the point-to-plane association as LOAM \cite{zhang2014loam}.


\subsection{Pose optimization}
\label{sec:pose_optimization}

We adopt the iEKF from FAST-LIO2 to optimize the pose.
The prediction step is implemented by the integration of IMU measurements from the latest optimized state $\overline{\mathbf{x}}_{k-1}$ along with the covariance matrix $\overline{\mathbf{P}}_{k-1}$.

For the residual computation, given a point $^\mathcal{W}\boldsymbol{p}_i$ in the world frame, the residual $z_i$ is calculated as:
\begin{equation}
        \mathbf{z}_i = \boldsymbol{n}_j(^\mathcal{W}\boldsymbol{p}_i - ^\mathcal{W}\boldsymbol{q}_j)
        \label{eq_z_i}
\end{equation}
where $\boldsymbol{n}_j$ is the normalized normal of the associated surfel or plane for $\boldsymbol{p}_i$, and $^\mathcal{W}\boldsymbol{q}_j$ is a point lying on the associated element.

Then, we denote the propagated state and covariance by $\widehat{\mathbf{x}}_{k}$ and $\widehat{\mathbf{P}}_{k}$ respectively. They represent the prior Gaussian distribution for the state.
By incorporating the prior distribution and the measurement models for point-to-surfel and point-to-plane associations from \Refeq{eq_z_i}, we obtain the maximum a-posterior estimate (MAP) as follows:
\begin{equation}
        \begin{aligned}
                \underset{\widetilde{\mathbf{x}}_k^\kappa}{min} & ( \|\mathbf{x}_{k}\boxminus \widehat{\mathbf{x}}_k\Vert^2_{\widehat{\mathbf{P}}_k}                                     
                                                                 +\sum_{i\in surfel} \|\mathbf{z}_i^\kappa + \mathbf{H}_i^\kappa \widetilde{\mathbf{x}}_k^\kappa \Vert_{\mathbf{R}_i}^2 \\
                                                                & +\sum_{j\in plane} \|\mathbf{z}_j^\kappa + \mathbf{H}_j^\kappa \widetilde{\mathbf{x}}_k^\kappa \Vert_{\mathbf{Q}_j}^2)
        \end{aligned}
        \label{eq_map}
\end{equation}
where $\boxminus$ computes the difference  between $\mathbf{x}_k$ and $\widehat{\mathbf{x}}_k$ in the local tangent space of $\mathbf{x}_k$,
$\widetilde{\mathbf{x}}_k^\kappa$ is the error of the $\kappa$-th iterate update at time $k$, $\mathbf{H}_i^\kappa$ and $\mathbf{H}_j^\kappa$ are Jacobian matrices with respect to $\widetilde{\mathbf{x}}_k^\kappa$, $\mathbf{R}_i$ and $\mathbf{Q}_j$ come from the raw measurement noise.
Compared with FAST-LIO2, we augment the MAP with point-to-surfel associations, which are the middle term of \Refeq{eq_map}.
The Kalman gain can be computed efficiently, with the computation load depending on the state dimension instead of the measurement dimension \cite{xu2021fast,xu2022fast}.

\subsection{Map Management}
\label{sec:map_management}
LOG-LIO uses an extended ikd-tree to manage the map.
The ikd-tree originally stores map points in both leaf nodes and internal nodes \cite{xu2022fast}.
In our extension, we additionally store a distribution in each node, and this distribution is maintained through voxels. 
Upon the first scan, we initialize the tree-structured map with a predetermined voxel resolution and associate the distribution of points within each voxel with the corresponding tree node. 
For subsequent points, if they fall within the same voxel as the nearest associated map point, we incrementally update the distribution within the voxel (\Refsec{sec:distribution_estimation}).
In cases where the points belong to a different voxel, we initialize a new tree node encompassing those points and the voxel and add it to the map.

To balance computational efficiency and accuracy, we limit the number of points added to each voxel by tuning the downsampling rate in pre-processing.
Additionally, we consider the distribution stabilizes once the directions of $\boldsymbol{n}_r$ and $\boldsymbol{e}_d$ converge.
The distribution is then fixed in the map. 



\section{EXPERIMENTAL RESULTS}
\label{sec:experimental_results}

\subsection{Implementation details}
\label{sec:implementation}
In data pre-processing, we set the downsampling grid to match the map's voxel size, ensuring that each voxel contains at most one point per frame. 
And we normalize the normal after downsamping.
Within the extended ikd-tree nodes, we maintain the point distribution, updating it once the voxel accumulates $\eta = 25$ points.
When the angle between $\boldsymbol{n}_r$ and $\boldsymbol{e}_d$ falls below 20 degrees, we consider the distribution stabilized, and both $\boldsymbol{n}_r$ and $\boldsymbol{e}_d$ are fixed and replaced by the average of their values.
For map voxels that accumulate $2\eta$ points without stabilization, we conclude that their distribution no longer requires updates and fix it.

\subsection{Experimental Settings}
The experiment focuses on the following two research questions:
\begin{itemize}
        \item
              Can Ring FALS estimates the normal of LiDAR points in real-time and accurately represents environmental information?
        \item
              Can LOG-LIO improve the accuracy of pose estimation by incorporating normal and distribution of points estimation?
\end{itemize}

\begin{figure*}[!ht]
        \centering
        \includegraphics[width=17cm]{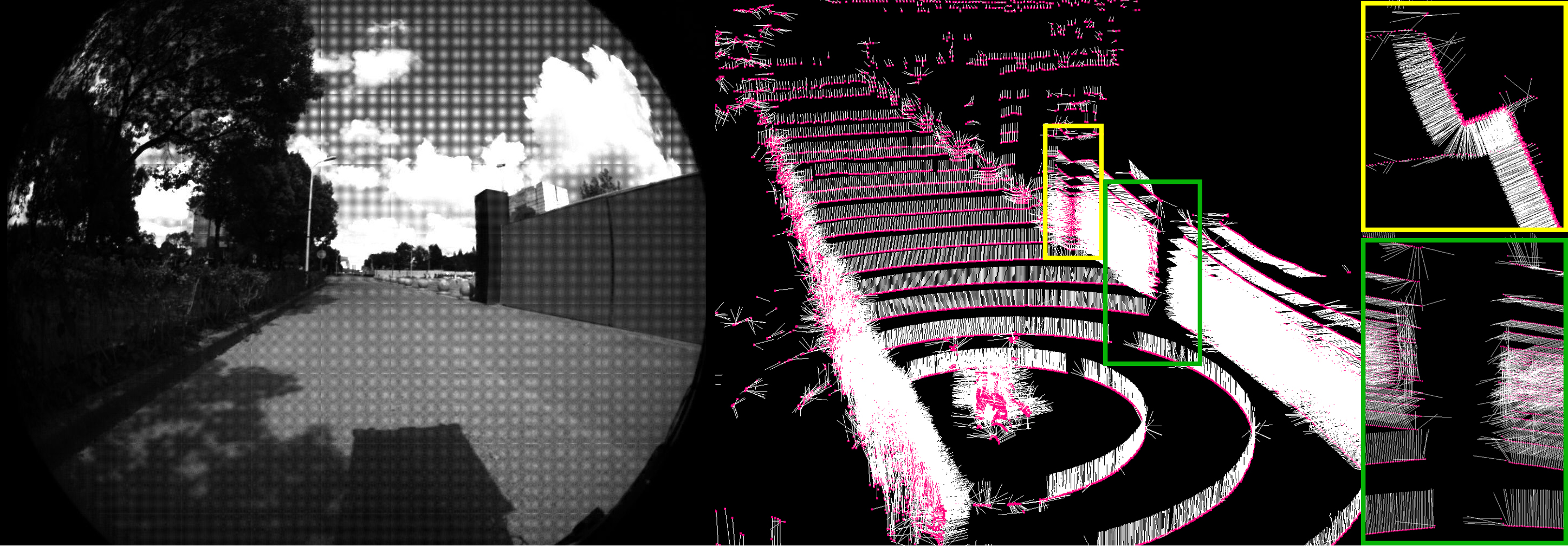}
        \caption{The starting position of the sequence \emph{gate03} of M2DGR dataset.
                The white lines represent normalized normals from Ring FALS estimation.}
        \label{fig_normal}
\end{figure*}
\begin{table*}[htb!]
        \caption{The Mean Running Time(ms) of Normal Estimation for A Single Scan to Certain LiDARs}
        \centering
        \begin{tabular}{c|c|c c c|c|c|c}
                \toprule
                            & \multirow{2}{*}{points} & \multicolumn{4}{c|}{Ring FALS} & \multicolumn{2}{c}{PCL}                                                               \\
                            &                         & projection                     & box-filtering           & smoothing & total          & single thread & OMP 10 threads \\
                \midrule
                Velodyne-32 & 57600                   & 2.045                          & 2.540                   & 3.199     & \textbf{7.784} & 79.811        & 26.355         \\
                Ouster-16   & 16384                   & 0.560                          & 1.221                   & 0.815     & \textbf{2.597} & 155.972       & 39.664         \\
                \bottomrule
        \end{tabular}
        \label{tab_ringfals_time}
\end{table*}
We conduct extensive experiments on the M2DGR\cite{yin2021m2dgr} and NTU VIRAL\cite{nguyen2022ntu} datasets, both of which include 9-axis IMU measurements and ground truth trajectories. 
The M2DGR dataset collects data on a ground platform equipped with Velodyne-32 LiDAR and captured in diverse indoor and outdoor scenarios with ground truth trajectories obtained from laser 3D tracking, motion capture, and RTK receivers.
The NTU VIRAL dataset collects data on an Unmanned Aerial Vehicle (UAV) platform with ground truth obtained by a laser-tracker total station with centimeter-level accuracy.
The horizontal Ouster 16-channel OS1 LiDAR and VectorNav VN100 IMU are used.
Compared with M2DGR, the LiDAR used by VIRAL has a sparser point cloud, making it more challenging to estimate poses in open areas.
In the experiments, the resolution of maps and new scan downsampling size are set to 0.4 m for M2DGR and 0.5 m for NTU VIRAL respectively.
Our workstation runs with Ubuntu 18.04, equipped with an Intel Core Xeon(R) Gold 6248R 3.00GHz processor and 32GB RAM.
\subsection{Evaluation of Normal Estimation}
\label{sec:expe_normal}

We conduct a comparative analysis of Ring FALS and PCL \cite{rusu20113d} normal estimation tools.
The implementation of PCL normal estimation is based on traditional least squares with the assistance of kdtree.
PCL also provides a parallel implementation using OpenMP to speed up the computation.
Note that normal smoothing is a time-consuming process for PCL, so only Ring FALS smoothness the normals.

To provide an intuitive evaluation, we visualize the normals at the starting position of sequence \emph{gate03} of the M2DGR dataset.
As shown in \Reffig{fig_normal}, the white lines represent the normalized normals estimated by Ring FALS. 
Notably, almost all ground point normals exhibit a vertical upward orientation.
With respect to the pillar in the yellow box, the normals at the corners transit smoothly with the normals on the adjacent sides.
Due to occlusion, points near fake edges within the green box fail to meet the neighborhood range similarity assumption, leading to inaccurate Ring FALS estimation.
However, these points with misestimated normals are a minority within the scan and are filtered out during visibility and consistency checks (\Refsec{sec:consistency_checks}).

\Reftab{tab_ringfals_time} shows the average processing times of normal estimation for a single LiDAR scan from the M2DGR and NTU VIRAL datasets, respectively.
The M2DGR dataset contains about 57,600 points per scan. 
Ring FALS demonstrates significantly reduced processing time compared to PCL, taking only one-tenth of the time, and it's four times faster than the OpenMP version. 
For NTU VIRAL, Ring FALS also achieves significantly shorter processing times compared to PCL, with or without OpenMP.

It is noteworthy that despite the Ouster-16 LiDAR having fewer points than Velodyne-32 in one scan, the consumption time increases.
This is due to the time-consuming kdtree neighborhood search in PCL's normal estimation. 
And it is influenced by the spatial structure of the kdtree, which, in turn, reflects the complexity of the environment.


\subsection{Evaluation of Odometry}
\label{sec:eva_odometry}

We compare LOG-LIO with two state-of-the-art LIO methods,  FAST-LIO2 \cite{xu2022fast} and LIO-SAM \cite{shan2020lio} without enabling loop closure. 
Accuracy assessment is based on the root-mean-square error (RMSE) of absolute trajectory error (ATE). 
We employ LOG-C, which only performs point-to-plane association, for ablation experiments.

\subsubsection{M2DGR Datasets}
Due to the instability of the RTK signal, the first 100 seconds and the last 100 seconds of \emph{street07} and \emph{street10} are discarded in the experiment.

\Reftab{tab_m2dgr} reports the quantitative results.
Notably, LOG-LIO, LOG-C, and FAST-LIO2 show comparable accuracy in indoor scenes, such as \emph{doors} and \emph{halls}, outperforming LIO-SAM in most cases.
This is due to the abundance of planar features in indoor scenes, which results in more map points forming true planes. 
Consequently, the point-to-plane data association provides effective constraints for pose estimation.
Conversely, the point-to-line data association of LIO-SAM may become less reliable, especially when errors accumulate.

\begin{figure*}[!ht]
        \centering
        \includegraphics[width=17cm]{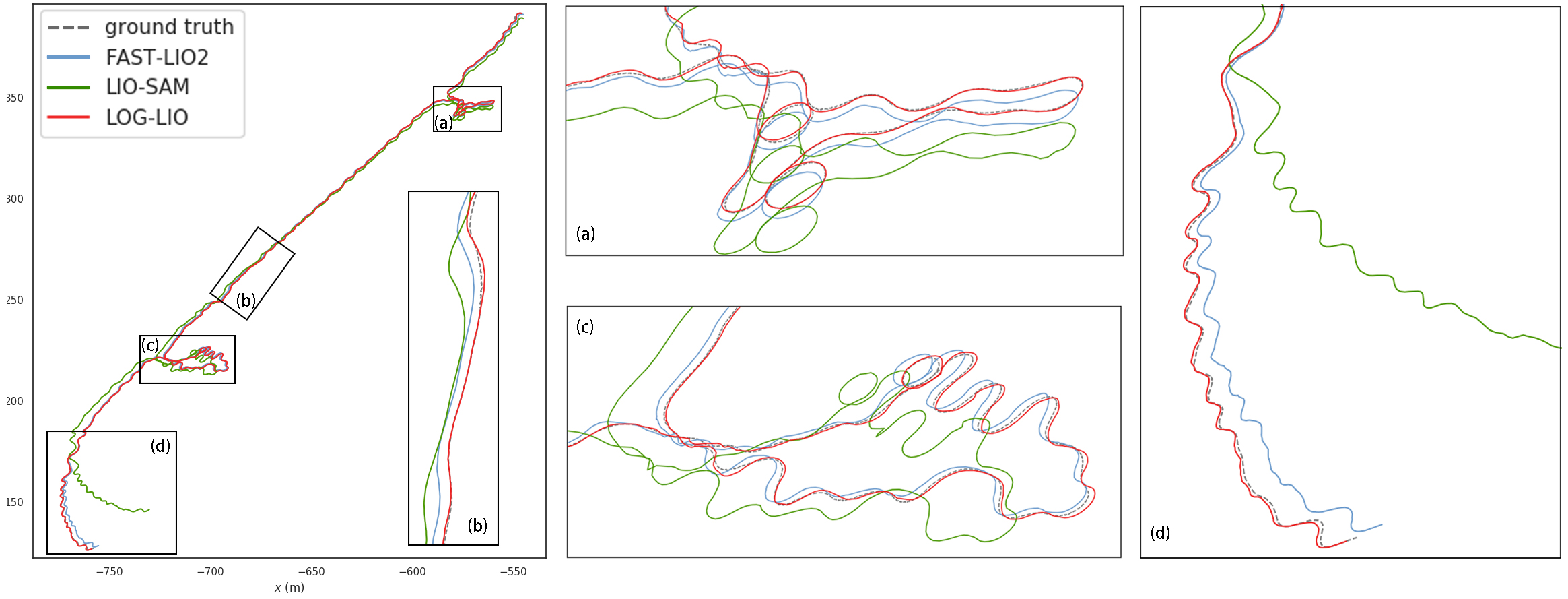}
        \caption{Localization estimates in sequence \emph{street10} of the M2DGR dataset.
                 The zoomed in image of the colored boxes corresponds to the boxes of the same color in the trajectory.}
        \label{fig_street10}
\end{figure*}

In outdoor sequences, \ie{ \emph{gate},  \emph{street}}, map points are relatively sparse compared to indoor scenes. 
This is especially evident in the \emph{street} sequences, where the robot moves on wide campus roads at night.
\Reffig{fig_street10} shows the trajectories of sequence \emph{street10} for qualitative comparison.
The competitive results of LOG-LIO suggest that efficient and accurate estimation of local geometric information exhibits great potential in reducing the error of LIO system.


Overall, when employing only point-to-plane association, LOG-C demonstrates a slight improvement in average accuracy when compared to FAST-LIO2.
However, with the implementation of our proposed hierarchical data association and map management scheme, LOG-LIO consistently achieves the lowest mean error and gets the best results in 10 out of the 21 sequences.

\begin{table}[htb!]
        \caption{The Translation RMSE(m) Results of Pose Estimation Comparison on the M2DGR Dataset}
        \centering
        \begin{tabular}{c|c|c|c|c|c}
                \toprule
                \scriptsize{Seq} & \scriptsize{duration(s)} & \scriptsize{LOG-LIO} & \scriptsize{LOG-C} & \scriptsize{FAST-LIO2} & \scriptsize{LIO-SAM} \\
                \midrule
                gate01           & 172                      & 0.097                & \textbf{0.085}     & \underline{0.091}      & 0.122                \\
                gate02           & 327                      & \textbf{0.270}       & \underline{0.277}  & 0.279                  & 0.288                \\
                gate03           & 283                      & \textbf{0.085}       & \textbf{0.085}     & 0.109                  & \underline{0.095}    \\
                walk01           & 291                      & \underline{0.078}    & \textbf{0.077}     & 0.112                  & 0.080                \\
                door01           & 461                      & \underline{0.251}    & \textbf{0.249}     & 0.271                  & 0.269                \\
                door02           & 127                      & \textbf{0.172}       & \underline{0.177}  & 0.200                  & 0.180                \\
                street01         & 1028                     & \textbf{0.246}       & \underline{0.294}  & 0.329                  & 0.559                \\
                street02         & 1227                     & \underline{2.448}    & \textbf{2.228}     & 2.754                  & 3.320                \\
                street03         & 354                      & \textbf{0.097}       & 0.103              & 0.106                  & \underline{0.102}    \\
                street04         & 858                      & \textbf{0.485}       & 0.565              & \underline{0.552}      & 1.009                \\
                street05         & 469                      & \underline{0.331}    & \textbf{0.287}     & 0.377                  & 0.407                \\
                street06         & 494                      & \underline{0.342}    & 0.391              & 0.434                  & \textbf{0.332}       \\
                street07         & 829                      & \underline{2.916}    & 3.646              & 3.512                  & \textbf{1.614}       \\
                street08         & 491                      & \textbf{0.130}       & \underline{0.146}  & 0.170                  & 0.161                \\
                street09         & 907                      & \underline{3.164}    & 3.754              & 3.648                  & \textbf{2.657}       \\
                street10         & 810                      & \textbf{0.388}       & 0.977              & \underline{0.956}      & 8.560                \\
                hall01           & 351                      & \textbf{0.256}       & \textbf{0.256}     & \underline{0.258}      & 0.281                \\
                hall02           & 128                      & \underline{0.274}    & \textbf{0.272}     & \underline{0.274}      & 0.285                \\
                hall03           & 164                      & \underline{0.345}    & 0.359              & \textbf{0.343}         & 0.579                \\
                hall04           & 181                      & \textbf{0.944}       & \textbf{0.944}     & \underline{0.952}      & 1.076                \\
                hall05           & 402                      & \underline{1.045}    & 1.046              & 1.049                  & \textbf{1.015}       \\
                \midrule
                mean             &                          & \textbf{0.684}       & \underline{0.772}  & 0.799                  & 1.095                \\
                \bottomrule
        \end{tabular}
        \label{tab_m2dgr}
        \begin{tablenotes}
                \footnotesize
                \item The best and second-best results are bolded and underlined respectively.
        \end{tablenotes}
\end{table}

\subsubsection{NTU VIRAL Datasets}
\label{sec_exp_viral}
As depicted in \Reftab{tab_viral}, LOG-LIO demonstrates the best performance in most sequences, closely followed by LOG-C. 
LIO-SAM achieves the best results on several sequences but fails in half of the dataset.

In the sequences \emph{nya} and \emph{tnp}, where the drone traverses indoors, LOG-LIO, LOG-C and FAST-LIO2 exhibit similar errors.
This behavior is consistent with what we observed in M2DGR, where the presence of numerous planes in confined spaces can effectively constrain the point-to-plane association.
In the \emph{eee}, \emph{sbs}, and \emph{rtp} outdoor scenes amidst buildings, LOG-LIO, LOG-C, and FAST-LIO2 yield highly similar trajectories due to effective constraints imposed by the plane structure.
However, in the \emph{spms} sequences, where the drone departs from an area surrounded by buildings and ascends to higher altitudes, the sparse LiDAR points lead to limited map overlap. 
This can potentially result in registration errors when performing point-to-plane data association.
LOG-LIO addresses this challenge by performing point association with corresponding voxels. 
The accurate local geometric information within these voxels help mitigate registration errors, ultimately resulting in more precise trajectories.

It is worth noting that in practice, trajectory error in the LIO system should consider various factors such as map resolution, IMU noise, and etc. 
And we focus on factors closely tied to our contributions while keeping the other parameters fixed in this paper.

\begin{table}[htb!]
        \caption{The Translation RMSE(m) Results of Pose Estimation Comparison on the NTU VIRAL Dataset}
        \centering
        \begin{tabular}{c|c|c|c|c|c}
                \toprule
                \scriptsize{Seq} & \scriptsize{duration(s)} & \scriptsize{LOG-LIO} & \scriptsize{LOG-C} & \scriptsize{FAST-LIO2} & \scriptsize{LIO-SAM} \\
                \midrule
                eee\_01              & 399      & 0.084                & \underline{0.082}      & 0.084                      & \textbf{0.049}           \\
                eee\_02              & 321      & \underline{0.072}    & \underline{0.072}      & 0.073                      & \textbf{0.051}           \\
                eee\_03              & 181      & \underline{0.112}    & 0.114                  & 0.113                      & \textbf{0.081}           \\
                nya\_01              & 396      & 0.081                & \underline{0.080}      & \textbf{0.075}             & 0.174                    \\
                nya\_02              & 428      & 0.111                & 0.111                  & \underline{0.109}          & \textbf{0.085}           \\
                nya\_03              & 411      & \textbf{0.120}       & \underline{0.121}      & \underline{0.121}          & 0.249                    \\
                sbs\_01              & 354      & \textbf{0.094}       & \underline{0.095}      & 0.097                      & x                        \\
                sbs\_02              & 373      & \underline{0.085}    & 0.086                  & \textbf{0.080}             & \textbf{0.080}           \\
                sbs\_03              & 389      & \underline{0.085}    & \textbf{0.083}         & \textbf{0.083}             & x                        \\
                rtp\_01              & 482      & \textbf{0.191}       & 0.230                  & \underline{0.209}          & x                        \\
                rtp\_02              & 453      & \underline{0.147}    & 0.153                  & 0.163                      & \textbf{0.117}           \\
                rtp\_03              & 355      & 0.180                & 0.181                  & \underline{0.170}          & \textbf{0.113}           \\
                tnp\_01              & 583      & \textbf{0.093}       & 0.095                  & \underline{0.094}          & x                        \\
                tnp\_02              & 457      & 0.075                & \textbf{0.058}         & \underline{0.071}          & x                        \\
                tnp\_03              & 407      & 0.084                & \textbf{0.079}         & \underline{0.080}          & x                        \\
                spms\_01             & 446      & \textbf{1.450}       & \underline{1.670}      & 1.818                      & x                        \\
                spms\_02             & 398      & \textbf{2.196}       & \underline{2.748}      & 3.378                      & x                        \\
                spms\_03             & 386      & \textbf{0.685}       & \underline{0.766}      & 0.793                      & x                        \\
                \midrule
                mean                 &          & \textbf{0.330}       & \underline{0.379}      & 0.423                      & x                        \\
                \bottomrule
        \end{tabular}
        \label{tab_viral}
        \begin{tablenotes}
                \footnotesize
                \item The best and second-best results are bolded and underlined respectively.
        \end{tablenotes}
\end{table}

\subsubsection{Processing Time Evaluation}
\begin{table*}[htb!]
        \caption{The Average Time Consumption(ms) of Each Sequence in The Experiments}
        \centering
        \begin{tabular}{c|c|c|c|c|c|c|c|c|c|c|c|c}
                \toprule
                {}        & \multicolumn{5}{c|}{M2DGR} & \multicolumn{6}{c|}{NTU VIRAL} &                                                                                                  \\
                {}        & gate                       & walk                           & door   & street & hall   & eee    & nya    & sbs    & rtp    & tnp    & spms   & mean            \\
                \midrule
                LOG-LIO   & 46.563                     & 45.964                         & 24.083 & 42.480 & 24.903 & 20.991 & 18.027 & 17.997 & 25.388 & 19.412 & 23.348 & 28.101          \\
                FAST-LIO2 & 31.378                     & 32.276                         & 14.754 & 28.970 & 15.509 & 15.705 & 12.476 & 12.785 & 21.006 & 13.340 & 17.106 & \textbf{20.523} \\
                \bottomrule
        \end{tabular}
        \label{tab_lio_time}
\end{table*}

We perform statistical analysis on the time consumption of LOG-LIO and FAST-LIO2 in each sequence, as shown in \Reftab{tab_lio_time}.
It is observed that the average processing time per scan of LOG-LIO is slightly longer than that of FAST-LIO2, which is mainly due to the Ring FALS normal estimation and incremental point distribution maintenance within map voxels.
Despite LOG-LIO exhibiting an additional average time consumption of 8 ms than FAST-LIO2, it still meets real-time requirements.
This performance difference should be considered in light of the number of points and the complexity of the environment.



\section{CONCLUSION AND FUTURE WORK}
\label{sec:conclusion_and_future_work}
This paper introduces LOG-LIO, an online LiDAR-inertial odometry method that estimates normal and distribution of points for local geometric information in real time. 
To improve normal estimation efficiency for LiDAR scans, 
we introduce Ring FALS, an efficient normal estimator that pre-records structural information and uses range data for normal estimation.
In LOG-LIO, we manage the map using an extended ikd-tree, incrementally maintaining normal and point distribution within map voxels.
We employ a hierarchical data association scheme for accurate constraints, resulting in precise pose estimation.
Experimental results show that LOG-LIO is competitive with state-of-the-art LIO systems in various environments.

For future research, we intend to incorporate dynamic noise removal and loop closure to enhance stability in dynamic environments and ensure long-term operation.
\bibliographystyle{IEEEtran}
\bibliography{IEEEabrv, paper}

\end{document}